\documentclass[pdflatex,sn-mathphys-num]{sn-jnl}

\usepackage{graphicx}%
\usepackage{multirow}%
\usepackage{amsmath,amssymb,amsfonts}%
\usepackage{amsthm}%
\usepackage{mathrsfs}%
\usepackage[title]{appendix}%
\usepackage{xcolor}%
\usepackage{textcomp}%
\usepackage{manyfoot}%
\usepackage{booktabs}%
\usepackage{listings}%

\usepackage{hyperref}       
\usepackage{url}            
\usepackage{booktabs}       
\usepackage{amsfonts}       
\usepackage{nicefrac}       
\usepackage{microtype}      
\usepackage{lipsum}
\usepackage{caption}
\usepackage{multicol}
\usepackage{multirow}
\usepackage{amsmath,amssymb}
\usepackage{wrapfig}
\usepackage[ruled,vlined,linesnumbered]{algorithm2e}

\usepackage{algpseudocode}
\usepackage{amsmath}
\usepackage{multirow}
\usepackage{subcaption}
\usepackage{float}     
\usepackage[numbers]{natbib}
 \usepackage{graphicx}
\usepackage{amssymb}
\usepackage{tcolorbox}
\newtcolorbox[number within=section]{promptbox}[2][]{%
  colback=blue!5,
  colframe=blue!40,
  coltitle=black,
  fonttitle=\bfseries,
  title=Prompt~\thetcbcounter: #2,
  #1
}
\usepackage{fancyhdr}       
\usepackage{graphicx}       
\graphicspath{{media/}}     

\newcommand\mpnetmodel{\texttt{all-mpnet-base-v2}}
\newcommand\minilmmodel{\texttt{all-MiniLM-L6-v2}}
\newcommand\simcsemmodel{\texttt{sup-simcse-roberta-large}}

\newcommand\test{\texttt{MedAESQA}}
\newcommand\val{\texttt{BioGen2024-val}}
\newcommand\train{\texttt{BioGen2024-train}}
\newcommand\bioace{\texttt{BioACE}}

\begin{document}

\title[BioACE: An Automated Framework for Biomedical Answer and Citation Evaluations
]{BioACE: An Automated Framework for Biomedical Answer and Citation Evaluations
}


\author*{\fnm{Deepak} \sur{Gupta}}\email{deepak.gupta@nih.gov}

\author{\fnm{Davis} \sur{Bartels}}\email{davis.bartels@nih.gov}

\author{\fnm{Dina} \sur{Demner-Fushman}}\email{ddemner@mail.nih.gov}

\affil{\orgdiv{Division of Intramural Research}, \orgname{National Library of Medicine
} \\ \orgaddress{\street{8600 Rockville Pike}, \city{Bethesda}, \postcode{20894}, \state{MD}, \country{USA}}}


\abstract{With the increasing use of large language models (LLMs) for generating answers to biomedical questions, it is crucial to evaluate the quality of the generated answers and the references provided to support the facts in the generated answers. Evaluation of text generated by LLMs remains a challenge for question answering, retrieval-augmented generation (RAG), summarization, and many other natural language processing tasks in the biomedical domain, due to the requirements of expert assessment to verify consistency with the scientific literature and complex medical terminology. In this work, we propose \bioace{}, an automated framework for evaluating biomedical answers and citations against the facts stated in the answers. The proposed \bioace{} framework considers multiple aspects, including completeness, correctness, precision, and recall, in relation to the ground-truth nuggets for answer evaluation. We developed automated approaches to evaluate each of the aforementioned aspects and performed extensive experiments to assess and analyze their correlation with human evaluations. In addition, we considered multiple existing approaches, such as natural language inference (NLI) and pre-trained language models and LLMs, to evaluate the quality of evidence provided to support the generated answers in the form of citations into biomedical literature. With the detailed experiments and analysis, we provide the best approaches for biomedical answer and citation evaluation as a part of \bioace{} (\url{https://github.com/deepaknlp/BioACE}) evaluation package.  }




\maketitle

\section{Introduction} \label{sec:intro}

Evaluation of answers to clinical questions generated by AI models remains a manual task as no reliable automated approaches that approximate human judgments exist. In addition to being expensive and time-consuming, the size of data manageable by human annotators might not be sufficient to detect the differences in model performance~\cite{singhal2025toward}. The evaluation bottleneck must be eliminated to enable the use of AI models in practical applications, e.g., providing information to clinicians and patients, as well as developing better systems. For the latter, the automated metrics only need to agree with the direction of human judgments, i.e., reliably inform the researchers if one of the systems performs a given task better than the other, or if the changes to the system improved the results. Conversely, the use of systems in real-life applications requires confidence in each answer along several axes, such as factuality/correctness, completeness, succinctness, grammaticality, coherency/consistency, and fluency. 

As grammaticality, cohesion, and fluency of the answers generated by Large Language Models (LLMs) resemble human~\cite{kumar2024deep}, other evaluation aspects, particularly factuality, become more of a concern. Approaches to improving the factuality of the generated answer are converging on providing evidence to support each statement with references into reliable sources provided to the models before generation (Retrieval Augmented Generation) or found retroactively to support the generated answer. Evaluation of the support provided by the referenced documents was an intrinsic part of the large-scale question answering evaluations in the past. In the Text Retrieval Conference (TREC) evaluations, the systems provided each factual answer along with a document identifier, and the answers were (manually) judged to be correct only if the text submitted by the system answered the question and the document supported the answer~\cite{ellen2001overview}. 
The first generation of automated approaches to the evaluation of the answer text relied heavily on the lexical overlap of the reference standards and system-generated texts.. While showing good correlations with human judgments for the answers extracted from relevant documents, the metrics were criticized for their inability to capture paraphrases, synonymy, and meaning of the text~\cite{lin2006methods}. These shortcomings became more critical with the advent of generative models. Efforts to overcome the shortcomings of measuring lexical similarity and rely more on semantics include: augmenting the lexicon with embeddings~\cite{m2020holms}, natural language inference (NLI)~\cite{chen-etal-2021-nli-models}, matching answers on key facts ('nuggets')~\cite{pradeep2025great}, and using LLMs as agents that evaluate the answers~\cite{chanchateval}, often using a question answering approach~\cite{zhong-etal-2022-towards}. 

In this work, we introduce \bioace{} an evaluation framework for assessing the answer and citations against the stated facts in the answers.  For answer evaluation, we propose diverse aspects (completeness, correctness, precision, and recall against the ground-truth answer) to evaluate the generated answer. In particular, we analyzed the performance of the nuggetization approach for nugget-based comparison between reference and generated answers, leveraging Llama for automated nugget extraction and measuring similarity between the nuggets extracted from the automated responses and the reference answers. For completeness, we analyzed the performance of various pretrained and large language models. Similar to completeness, we evaluated the results using language models and the semantic similarity between the answer and relevant documents to assess the answer's correctness.  Extensive experiments are performed for each answer evaluation aspect to determine the best automated approach that yields the highest correlation with human evaluation. We also compared the best method, which yields the highest correlations, with existing automated approaches. We performed all experiments on the BioGen collection \cite{gupta2024overview} that consists of sixty-five (65) questions. We consider the first 40 questions as the test dataset, which is the same as the \test{} \cite{gupta2025dataset} dataset. The following five questions are considered the validation dataset, which we refer to as the \val{} set. The last 20 questions serve as the training set, which we refer to as \train{}. 
Detailed experiments and analyses are also performed to determine the best approach for automatically assessing citation quality. In particular, we analyzed existing methods for comparing answer assertions with cited documents and conducted experiments across a variety of setups to determine the capability of large language models for the citation evaluation task. Similar to the answer evaluation, we compared the best citation evaluation method with the existing techniques and drew our findings. 


We summarize the contributions of this work as follows:
\begin{enumerate}
    \item We propose automated approaches for evaluating answers to biomedical questions. The proposed approaches consider multiple aspects of the answer to determine its quality, such as correctness, completeness, and the coverage of facts. The facts take the form of answer nuggets both in the generated and the human-annotated answer. The proposed approaches are compared with a variety of models, ranging from standard machine learning models (SVM, Logistic regression) to the latest LLM-based models (zero-shot and fine-tuning approaches for Llama, Mistral, Qwen, etc.). 
    \item We experimented with a variety of approaches and settings to determine the best model that can assess whether the provided citation into literature supports the facts stated in the answer. These approaches are compared and analyzed on the human-annotated \test{} benchmark dataset to evaluate their effectiveness for the citation evaluation task. 
    \item Based on our extensive experiments with the answer and citation evaluation considering multiple aspects, settings, and models, we propose \bioace{} an answer and citation evaluation framework for biomedical question answering. This automatic framework includes the best models, parameters, and settings that obtained the highest performance on \test{} and shows strong correlation with the human annotation. 
\end{enumerate}

\section{Results}
The overarching goal of this study is to analyze the effectiveness of automated evaluation approaches for biomedical answers and their citations, and to assess their alignment with human judgments. Before assembling the best approaches to evaluating multiple aspects of the answers and integrating evidence into an evaluation framework, we present the evaluation results for each component. 
\subsection{Answer Evaluation} \label{sec:answer_eval}
For answer evaluation, we experimented with multiple aspects, and  the results for each of those aspects are provided below:
\subsubsection{Nugget Precision and Recall}
The detailed results in comparing the ground-truth and system-generated nuggets of the answer are shown in Table \ref{tab:res-precision_recall}. We found that the embedding model \simcsemmodel{} outperforms (precision: $44.68$, recall: $58.39$) the \minilmmodel{} and \mpnetmodel{}. 
We found the optimal probability threshold for \minilmmodel{} as $0.6267$, \simcsemmodel{} as $0.6035$, and \mpnetmodel{} as $0.6211$. We recorded the objective(s) value of F1-score (average similarity) to be $63.45$ ($65.25$), $55.72$ ($73.36$) and $62.68$ ($67.98$) for the \minilmmodel{}, \simcsemmodel{} and  \mpnetmodel{}, respectively. 

\begin{table}[h]
\centering
\begin{minipage}{0.75\linewidth}
\centering
\resizebox{\columnwidth}{!}{%
\begin{tabular}{llll}
\hline
\multicolumn{1}{c}{\textbf{\begin{tabular}[c]{@{}c@{}}Embedding\\ Model\end{tabular}}} & \textbf{Precsion} & \textbf{Recall} & \textbf{F-Score} \\ \hline \hline
all-MiniLM-L6-v2         & 36.43 & 47.74 & 41.33 \\ 
all-mpnet-base-v2        & 37.57 & 49.26 & 42.63 \\ 
sup-simcse-roberta-large & 44.68 & 58.39 & 50.62 \\ \hline \hline
\end{tabular}%
}
\caption{Performance comparison of the different models on comparing the ground-truth and generated answer nuggets in terms of precision, recall, and F1-scores.}
\label{tab:res-precision_recall}
\end{minipage}%
\hfill
\end{table}

\subsubsection{Completeness} 
The detailed performance of the PLMs and LLMs is demonstrated in Table \ref{tab:res-completeness}. The PLMs were fine-tuned on the \train{} dataset and the best model parameters are based on the performance on the \val{} dataset. The best model is used to evaluate the performance on the \test{} dataset. RoBERTa$_{\text{Large}}$ outperformed the other three pre-trained language models, achieving the highest weighted F1-score ($75.37$) across the individual classes (\textit{Required}, \textit{Unnecessary}, \textit{Borderline}, and \textit{Inappropriate}). Most of the PLMs achieved a competitive precision score, but RoBERTa$_{\text{Large}}$ recorded the highest recall score of $74.91$, while its counterpart, BERT$_{\text{Large}}$, achieved a recall score of $62.57$. We conducted detailed experiments using five different LLMs in two distinct settings: zero-shot and fine-tuned. In the zero-shot setting, we observe that Llama-3.3-70B-Instruct model achieves the highest weighted F1-score of $76.20$, while in the same Llama family model, Llama-3-8B-Instruct achieved only an F1-score of $63.70$. Among all the LLMs, Mistral-7B-Instruct-v0.3 achieved the lowest F1-score of $63.41$. In the fine-tuned setting of the LLMs Llama-3-8B-Instruct, achieved the highest F1-score of $72.67$. We observe that the performance of the fine-tuned Llama-3.3-70B-Instruct and Qwen3-8B decreases from $76.2$ to $63.7$ and from $64.89$ to $27.79$, respectively, compared to the zero-shot setting. Among all the LLMs, Mistral-7B-Instruct-v0.3 achieved the highest increase of 5.15 points in F1-score compared to its zero-shot setting. In both PLMs and LLMs experiments, Llama-3.3-70B-Instruct achieved the highest F1-score of $76.20$ in the zero-shot setting.
\begin{table}[]
\centering
\resizebox{\columnwidth}{!}{%
\begin{tabular}{lllll}
\hline
\textbf{Model} & \textbf{Setting} & \textbf{Precision} & \textbf{Recall} & \textbf{F1-Score} \\ \hline \hline
BERT$_{\text{Base}}$                   & Fine-tuned & 76.96 & 67.79 & 70.16 \\
BERT$_{\text{Large}}$      & Fine-tuned & 76.62 & 62.57 & 66.74 \\ 
RoBERTa$_{\text{Base}}$            & Fine-tuned & 76    & 65.9  & 68.64 \\ 
RoBERTa$_{\text{Large}}$           & Fine-tuned & 77.88 & 74.91 & 75.37 \\  \hline
Llama-3.3-70B-Instruct   & Zero-shot  & 80.72 & 73.26 & 76.2  \\ 
Llama-3.3-70B-Instruct   & Fine-tuned & 78.33 & 55.71 & 63.7  \\ 
Llama-3-8B-Instruct & Zero-shot  & 72.93 & 68.46 & 70.49 \\ 
Llama-3-8B-Instruct & Fine-tuned & 71.54 & 79.56 & 72.67 \\ 
Qwen3-14B                      & Zero-shot  & 76.53 & 70.03 & 70.5  \\ 
Qwen3-14B                     & Fine-tuned & 73.26 & 71.96 & 71.62 \\ 
Qwen3-8B                      & Zero-shot  & 74.35 & 59.74 & 64.89 \\ 
Qwen3-8B                       & Fine-tuned & 78.19 & 17.45 & 27.79 \\ 
Mistral-7B-Instruct-v0.3  & Zero-shot  & 69.02 & 60.07 & 63.41 \\ 
Mistral-7B-Instruct-v0.3  & Fine-tuned & 80.44 & 71.27 & 68.56 \\ \hline \hline
\end{tabular}%
}
\caption{Performance of the different PLMs and LLMs on the task of answer completeness. The reported results are the weighted precision, recall, and F1-score. }
\label{tab:res-completeness}
\end{table}

\subsubsection{Correctness}
Under the \textit{classification-based approach}, we have outlined the performance of various models, ranging from classical models to fine-tuned LLMs, in Table \ref{tab:res-answer_correctness}. We evaluated the performance of these models in terms of precision, recall, F1-score, and area under the curve for the binary classification task of assigning an answer sentence as correct or incorrect. For the classical model, SVM achieves the best F1-score of $79.15$ compared to its counterpart, the logistic regression model ($75.70$). We extend the experiments with the PLMs and analyze most of the PLMs that performed competitively, achieving F-scores in the range of $97.31$ (BERT$_{\text{Large}}$) to $97.65$ (RoBERTa$_{\text{Large}}$). With the LLMs, we conducted experiments in two settings: zero-shot and fine-tuning. For the zero-shot setting, Mistral-7B-Instruct-v0.3 achieves the highest precision score of $67.23$ while Llama-3-8B-Instruct recorded the lowest precision score of $28.77$. Qwen3-8B achieved balanced precision and recall scores of $58.77$ and $57.89$, respectively, leading to the best-performing model in the zero-shot setting. Under the fine-tuning setup, we observed no improvement over the zero-shot setup except for the Llama-3-8B-Instruct model. The Llama-3-8B-Instruct model achieves a gain in terms of F1-score, increasing from $32.4$ to $33.25$.  Qwen3-8B remains the best-performing model with the highest F1-score of $48.03$. Fine-tuning could not benefit the models, highlighting the need for model-dependent prompt and training configurations (such as objective and hyperparameters) to improve performance.  

For the \textit{semantic similarity-based approach}, we found that the average positive scores for document-answer are much higher than the NLI model probability for the \textit{support} score. Similar observations are made for the evidence-answer pair. The NLI model shows low probability for the negative pair, and on the contrary, the sup-simcse-roberta-large model shows high cosine similarity for the negative pair. We also compute the metric accuracy, which refers to the proportion of samples in which the score assigned to a positive pair exceeds that of its corresponding negative pair, relative to the total number of samples. We found that the accuracy of cosine similarity for both document and evidence settings is higher than that of the corresponding NLI-based scoring. The detailed results are depicted in Figure \ref{fig:ac-similarity-analysis}.
\begin{table}[]
\centering
\resizebox{\columnwidth}{!}{%
\begin{tabular}{llcccc}
\hline
\textbf{Model Type}   & \textbf{Model Name}       & \textbf{Precision} & \textbf{Recall} & \textbf{F1-Score} & \textbf{AUC} \\ \hline \hline
\multirow{2}{*}{Classical}       & SVM                     & 81.08 & 79.47 & 79.15 & 88.52 \\ 
                                 & Logistic Regression                          & 75.93 & 75.76 & 75.7  & 84.13 \\ \hline
\multirow{4}{*}{PLMs} & RoBERTa$_{\text{Large}}$   & 97.65              & 97.65           & 97.65             & 99.34        \\
                                 & RoBERTa$_{\text{Base}}$       & 97.48 & 97.47 & 97.47 & 99.26 \\ 
                                 & BERT$_{\text{Large}}$  & 97.35 & 97.32 & 97.31 & 99.08 \\  
                                 & BERT$_{\text{Base}}$              & 97.48 & 97.46 & 97.45 & 99.13 \\ \hline
\multirow{5}{*}{Zero-shot} & Qwen3-8B                        & 58.17 & 57.89 & 57.49 & NA\\ 
                                 & Llama-3.3-70B-Instruct          & 38.33 & 38.93 & 38.11 & NA \\ 
                                 & Llama-3-8B-Instruct        & 28.77 & 46.97 & 32.4  & NA \\
                                 & Mistral-7B-Instruct-v0.3        & 67.23 & 51.42 & 37.08 & NA \\ 
                                 & Qwen3-14B                       & 53.37 & 51.26 & 42.09 & NA \\ \hline
\multirow{5}{*}{Fine-tuned}  & Qwen3-8B                        & 52.95 & 52.05 & 48.03 & NA \\  
                                 & Llama-3.3-70B-Instruct         & 24.87 & 50.01 & 33.22    & NA \\  
                                 & Llama-3-8B-Instruct       & 58.21 & 50.01 & 33.25& NA\\  
                                 & Mistral-7B-Instruct-v0.3        & 49.55 & 49.99 & 33.79 & NA \\ 
                                 & Qwen3-14B                       & 63.11 & 52.73 & 40.94 & NA \\ \hline \hline
\end{tabular}%
}
\caption{Performance comparison of the different model types under multiple experimental settings for the answer correctness evaluation. The reported precision, recall, and F1-scores are macro-averaged over the correct and incorrect classes. Since LLM experiments are conducted in a generative manner without extracting the class probability, the AUC scores do not apply to these models.}
\label{tab:res-answer_correctness}

\end{table}

\subsection{Citation Evaluation}
We provide results for the citation evaluation task for several models with different architectures and with and without fine-tuning. We fine-tune the selected models on the \texttt{BioGen2024-train} dataset and choose the best model parameters based on the performance on the \texttt{BioGen2024-val} dataset. The transformer architecture models are fine-tuned using Low-Rank Adaptation (LoRA)~\cite{DBLP:journals/corr/abs-2106-09685}. For the models that produce scores, \texttt{alignscore}, \texttt{summacconv}, and \texttt{summaczs}, we fit their threshold to the training set to maximize F1-Score.

The results for the Answer Sentence-Document setting are as shown in table \ref{tab:res-binary} and table \ref{tab:res-ternary} representing the binary and ternary labeling, respectively. The results for the Answer Sentence-maxSimSentence Document setting are as shown in table \ref{tab:res-binary-maxSimSentence} and table \ref{tab:res-ternary-maxSimSentence} representing the binary and ternary labeling, respectively. Finally, the results for the Answer Nuggets-Document Nuggets setting are given in table \ref{tab:res-ternary-nuggets}, for which we only evaluated ternary labeling. The results were concentrated across all settings. However, transformer architectures tended to perform slightly higher. Fine-tuned and fitted models only yielded modest improvements over their base and unfitted counterparts. Our largest model, Lamma-3.3 performed the best, but that performance had low variance across our settings and prompts.

\begin{table}[]
\centering
\resizebox{0.9\columnwidth}{!}{%
\begin{tabular}{lllll}
\hline
\textbf{Model} & \textbf{Setting} & \textbf{Precision} & \textbf{Recall} & \textbf{F1-Score} \\ \hline \hline
Llama-3.3           & Base        & 76.65 & 76.64 & 76.64 \\
FLAN-T5             & Base        & 74.46 & 73.18 & 73.81 \\
FLAN-UL2            & Base        & 75.96 & 72.23 & 74.04 \\
alignscore          & Unfitted    & 75.23 & 53.78 & 62.70 \\
attrscore\_alpaca   & Base        & 75.67 & 52.04 & 61.62 \\
attrscore\_flan\_t5 & Base        & 74.18 & 75.62 & 74.89 \\
t5\_xxl\_true       & Base        & 77.56 & 57.34 & 65.92 \\
summacconv          & Unfitted    & 75.80 & 34.12 & 46.97 \\
summaczs            & Unfitted    & 70.09 & 59.73 & 64.49 \\
Llama-3.3           & Fine-tuned  & 78.12 & 77.85 & 77.98 \\
FLAN-T5             & Fine-tuned  & 75.89 & 74.56 & 75.22 \\
FLAN-UL2            & Fine-tuned  & 77.34 & 73.89 & 75.57 \\
alignscore          & Fitted      & 76.88 & 55.12 & 64.12 \\
attrscore\_alpaca   & Fine-tuned  & 77.05 & 53.21 & 63.02 \\
attrscore\_flan\_t5 & Fine-tuned  & 75.62 & 76.91 & 76.26 \\
t5\_xxl\_true       & Fine-tuned  & 79.01 & 58.97 & 67.42 \\
summacconv          & Fitted      & 77.03 & 35.84 & 48.88 \\
summaczs            & Fitted      & 71.88 & 60.95 & 65.93 \\ \hline \hline
\end{tabular}%
}
\caption{Performance of base and fine-tuned models when tasked with assigning binary labels to a claim sentence and a PubMed title and abstract.}
\label{tab:res-binary}
\end{table}

\begin{table}[]
\centering
\resizebox{0.9\columnwidth}{!}{%
\begin{tabular}{lllll}
\hline
\textbf{Model} & \textbf{Setting} & \textbf{Precision} & \textbf{Recall} & \textbf{F1-Score} \\ \hline \hline
Llama-3.3           & Base        & 77.23 & 77.05 & 77.14 \\
FLAN-T5             & Base        & 75.10 & 73.95 & 74.52 \\
FLAN-UL2            & Base        & 76.50 & 72.90 & 74.66 \\
alignscore          & Unfitted    & 75.80 & 54.60 & 63.52 \\
attrscore\_alpaca   & Base        & 76.20 & 51.80 & 61.72 \\
attrscore\_flan\_t5 & Base        & 74.50 & 76.00 & 75.24 \\
t5\_xxl\_true       & Base        & 78.10 & 57.90 & 66.46 \\
summacconv          & Unfitted    & 76.50 & 34.80 & 47.89 \\
summaczs            & Unfitted    & 71.00 & 60.50 & 65.29 \\
Llama-3.3           & Fine-tuned  & 78.80 & 78.50 & 78.65 \\
FLAN-T5             & Fine-tuned  & 76.50 & 74.90 & 75.69 \\
FLAN-UL2            & Fine-tuned  & 77.90 & 74.50 & 76.16 \\
alignscore          & Fitted      & 77.30 & 55.80 & 64.78 \\
attrscore\_alpaca   & Fine-tuned  & 77.80 & 52.90 & 62.88 \\
attrscore\_flan\_t5 & Fine-tuned  & 76.20 & 77.80 & 76.99 \\
t5\_xxl\_true       & Fine-tuned  & 79.80 & 59.50 & 68.10 \\
summacconv          & Fitted      & 77.80 & 36.10 & 49.24 \\
summaczs            & Fitted      & 72.60 & 61.80 & 66.73 \\\hline \hline
\end{tabular}%
}
\caption{Performance of base and fine-tuned models when tasked with assigning binary labels to a claim sentence and the sentence with the highest cosine similarity to the claim.}
\label{tab:res-binary-maxSimSentence}
\end{table}

\section{Discussion}

\begin{figure}
    \centering
    \includegraphics[width=\linewidth]{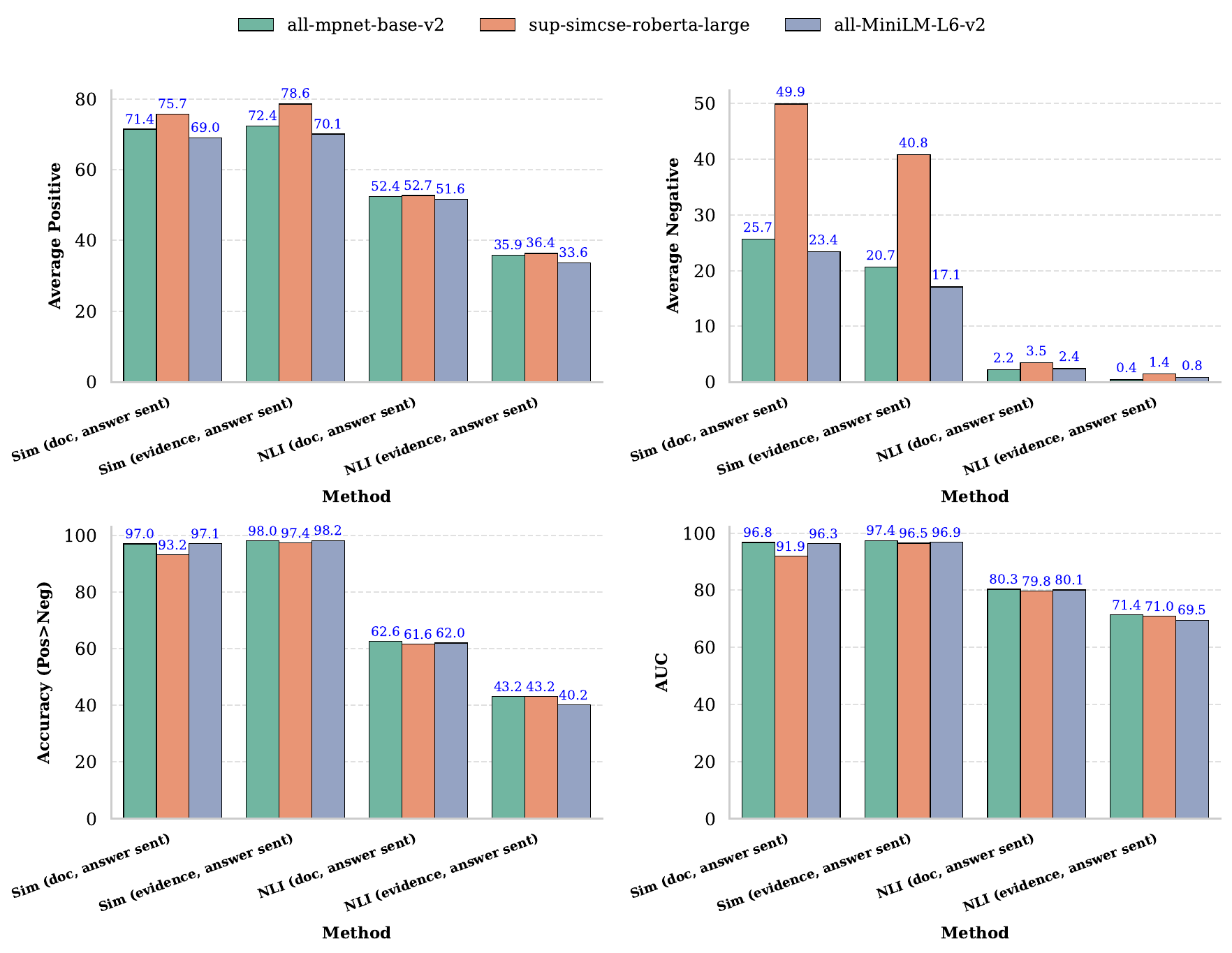}
    \caption{Answer correctness performance comparison of the embedding models, their derived cosine similarity, and the NLI model on multiple evaluation metrics.}
    \label{fig:ac-similarity-analysis}
\end{figure}
To analyze the effectiveness of the proposed answer evaluation metrics, we first analyze the usefulness of the recall metric by comparing it with the recall metric computed from the human-annotated answer sentences. We compare the proposed recall with the recall metric computed by following the R+B setting \cite{gupta2025dataset}, which involves including required or borderline answer sentences to cluster the system-generated answer sentences and compute the recall. The recall in this setup was formally defined as the ratio of clusters containing answer sentences to the total number of clusters. The answer sentences are encoded using the sup-simcse-roberta-large (SimCSE) and all-mpnet-base-v2 (ST) sentence embedding models. In particular, we rank the answer generation approaches (M1-M30) on the \test{}  dataset using the Recall (R+B)–SimCSE and Recall (R+B)–ST metrics. Thereafter, we also rank the answer generation approaches (M1-M30) based on the proposed nugget-based recall metric. With the rank of approaches using the human-evaluated recall metrics and the proposed recall metric, we compute the Spearman and Kendall Tau correlation coefficients and report the results (\textit{cf.} Table  \ref{tab:res-prec_recall_correlation}) for all three embedding models, which we experimented with using nugget-based precision-recall computation. We analyzed a strong correlation between the proposed and existing recall metrics and observed that the embedding model, all-MiniLM-L6-v2, yielded the highest correlation coefficients. The high correlation coefficients signify that the rankings of the system-generated answers are very consistent when using the proposed recall metric compared to the human-evaluated recall metrics. Based on these observations, the proposed automated nugget recall metric could be considered a potential recall metric for evaluating the recall of answer generation approaches.

To analyze the most effective model for the answer completeness task, we rank the answer generation approaches (M1-M30) on the \test{}  dataset using the Precision, Redundancy, and Harmfulness evaluation metrics introduced in \cite{gupta2025dataset}. The metrics are calculated based on the human-assessed relevance labels assigned to the answer sentences. We replaced the human-assessed relevance labels with the model-predicted labels and computed the metrics. We rank the answer generation approaches (M1-M30) based on both the model-predicted labels and the human-assessed relevance labels. With these two sets of system rankings, we compute the Spearman and Kendals Tau correlation coefficients and report the results (\textit{cf.} Table \ref{tab:res-completeness_correlations}) for all the PLMs and LLMs that we experimented with for answer completeness evaluation. We obtained the highest correlation with the Llama-3.3-70B-Instruct (zero-shot) model for the precision (answer completeness) metric, which is one of the key metrics for assessing answer completeness.  RoBERTa$_{\text{Large}}$ achieved the highest correlation for the Redundancy metric. The Llama-3.3-70B-Instruct (zero-shot) model achieved the highest correlation for Precision and a competitive correlation for the Redundancy and Harmfulness metrics, and could be considered a potential metric for evaluating answer completeness.  

In another analysis, we aim to assess the effectiveness of the trained answer correctness model on the \test{}  dataset. Given an answer sentence and its associated model-predicted citations, we tokenized the cited document. We prepared a list of three consecutive sentences and their corresponding answer sentence. We then use the trained answer correctness model to predict the correctness label for the pair of answer sentence and document. We consider the answer to be correct if its prediction for any grouped document sentence is deemed `\textit{correct}'. We experiment with two different settings. In one, we consider all the cited documents, regardless of their judgment of the answer sentence. In the other setting, we only consider the documents that were judged `supported' by the expert. We evaluate the correctness of each of the answer sentences in response to the question and average them over all the answer sentences. The detailed results with the best-performing PLMs are reported in Fig. \ref{fig:ac-participant-analysis}. The Supp(CS) metric achieved a high score in the range of 92-93, indicating that the trained correctness model exhibits high alignment in judging the correctness of the answer sentence with human judgment. In contrast, Supp(AS) achieved a score significantly lower than Supp(CS), as some of the documents were unsupported or not relevant, and the correctness model could not have found any grouped sentences in those documents to consider the correctness of the answer sentence. 

We extend this analysis to assess the effectiveness of answer correctness in practical scenarios where relevant documents are not readily available. Towards this, given the question of the \test{} dataset, we use the BM25 model to retrieve the top-1000 documents from the latest PubMed baseline index\footnote{\url{https://ftp.ncbi.nlm.nih.gov/pubmed/baseline/}}. Thereafter, we use the ranker model to rerank the BM-25 retrieved documents. Given the top retrieved documents of a given question, we aim to assess the correctness of each answer sentence in response to the given question.  We reported the performance (cf. Fig. \ref{fig:ac-top-doc-analysis}) by considering the top 10, 20, up to the top 100 documents, providing the Correctness at top-k (Correctness) scores. We observed that BERT$_{\text{Base}}$ and RoBERTa$_{\text{Large}}$ models quickly reach high correctness scores as the $k$ value increases. On the other hand, the BERT$_{\text{Large}}$ and RoBERTa$_{\text{Base}}$ models showed gradual improvement as $k$ increased and reached a plateau near $ k=90$. To further investigate this, we experimented with the NLI model and the SimCSE cosine similarity approach with a threshold of 0.75 as observed in Fig. \ref{fig:ac-similarity-analysis}. We reported the results in Fig. \ref{fig:ac-participant-top-doc}, where we found that SimCSE and NLI models also follow a similar pattern to RoBERTa$_{\text{Base}}$ and BERT$_{\text{Large}}$, reaching a plateau near $k=90$. 
We found that the BERT$_{\text{Large}}$ model consistently outperforms other PLMs across various settings for top-document retrieval and can be used as an answer-correctness evaluation model without requiring document-relevance assessment.

\subsection{Citation Evaluation}
We did not observe any significant differences among the models, settings, and prompts. 
Additionally, fine-tuning and fitting the models did not necessarily improve the results, contrary to conventional expectation, and even diminished results, in the case of the F1-Score for Llama-3.3 under the Answer Nuggets-Document Nuggets setting. This could be due to catastrophic interference~\cite{MCCLOSKEY1989109} or due to the relatively small size of the training set. For the scoring models, \texttt{alignscore}, \texttt{summacconv}, and \texttt{summaczs}, the optimal threshold tended to be as low as possible. This is likely because the data had more positive (attributable) cases, and therefore assigning more "attributable" labels yielded a higher F1-Score. We perform an analysis of rankings based on the automatically generated labels and compare this to the rankings from \test{} test dataset results on citation \cite{gupta2025dataset}. Correlation coefficients are computed in table \ref{tab:res-citation_correlation} and indicate low correlation with rankings based on the labels from medical informatics experts. Overall, performance leaves ample room for improvement, demonstrating the complexity of citation attribution, specifically in the medical domain.

\section{Methods}

\subsection{Answer Evaluation}
The quality of an answer to a biomedical question is comprehensively evaluated in a plan that covers a range of aspects. In particular, the answers are assessed in terms of \textit{Nugget precision}, \textit{Nugget recall}, \textit{completeness}, and \textit{correctness}. To compute the \textit{Nugget precision} and \textit{Nugget recall}, medical informatics experts have formulated the ground-truth answers and nuggets for each question following the guidelines discussed in \cite{gupta2025dataset}. We utilize the Llama 3.3 (70B) model to generate the nuggets from a system-generated answer using the approach and prompt discussed in \cite{bartels-etal-2025-large}. This section presents the models and settings explored for each of the answer evaluation aspects.
\subsubsection{Nugget Precision and Recall}
We begin the computations of precision and recall by extracting the sentence embeddings of the ground-truth and system-generated nuggets. To assess the role of embedding models comprehensively, we use a variety of embedding models (\minilmmodel{}\footnote{\url{https://huggingface.co/sentence-transformers/all-MiniLM-L6-v2}}, ~ \mpnetmodel{}\footnote{\url{https://huggingface.co/sentence-transformers/all-mpnet-base-v2}}, ~\simcsemmodel{}\footnote{\url{https://huggingface.co/princeton-nlp/sup-simcse-roberta-large}}). For each list of the ground-truth and system-generated nuggets, we created a similarity matrix $S$, where an item $S[i,j]$ represents the cosine similarity between the $i^{th}$ system-generated nugget and the $j^{th}$ ground-truth nugget. With the goal of finding the best cosine similarity threshold that optimizes the precision and recall between system-generated and ground-truth nuggets, we utilized the \train{} set that contains the system-generated and ground-truth nuggets for the question-answer pairs. Given the similarity matrix of each question-answer, we build a Bayesian Gaussian Mixture Model \cite{bishop2006pattern} (BGMM) to represent the observed distribution in the nugget similarities. We specified a maximum of two components (a pair of nuggets are semantically similar or dissimilar), allowing the model to represent the data with up to two latent clusters. The Gaussian mixture model parameters (mean and variance) are estimated using variational inference, which is an extension of expectation-maximization that maximizes a lower bound on model evidence (including priors) instead of data likelihood. Once the BGMM is trained, we use the model to predict the probability of a system-generated and ground-truth nugget being semantically similar. If they are predicted as similar, we align and match them; we follow the same procedure for each pair of system-generated and ground-truth nuggets. With the aligned nuggets, we compute the precision and recall of the answer nuggets. We use the same \train{} set that we use to train the BGMM model to tune the threshold, maximizing the multi-objectives (average F1-score, average nugget similarity, and average number of nugget alignments between the system-generated and ground-truth nuggets). We utilize Optuna\footnote{\url{https://optuna.readthedocs.io/en/stable/index.html}} to find the optimal probability threshold that maximizes the multi-objectives. 

\subsubsection{Completeness} To assess the completeness, we train a variety of classifiers on a question and each answer generated sentence, aiming to predict the label (\textit{Required}, \textit{Unnecessary}, \textit{Borderline}, or \textit{Inappropriate}). We use the pretrained language models: BERT$_{\text{Base}}$, BERT$_{\text{Large}}$, RoBERTa$_{\text{Base}}$, and RoBERTa$_{\text{Large}}$ to train the model on the \train{} dataset, tune the hyperparameters on the \val{} dataset, and evaluate the performance in terms of precision, recall, and F1-score on the \test{}  dataset. We also utilized large language models (Qwen3-8B, Qwen3-14B, Llama-3-8B-Instruct, Llama-3.3-70B-Instruct, and Mistral-7B-Instruct-v0.3) for this task, using them in both zero-shot and fine-tuned setups with the prompt specified in Figure \ref{fig:pr:completeness}. We use the low-rank adaptation \cite{hu2022lora} to fine-tune the LLMs on the \train{} dataset and report the classification performance on the \test{}  dataset. With the model-predicted classification label, we compute the completeness as the portion of the generated answer sentence that is classified as Required.

\subsubsection{Correctness} 
\paragraph{Classification Approach} We cast the assessment of the correctness of a generated answer sentence as a classification task in which an answer sentence and the supporting evidence are provided to a classifier to label the pair as correct or incorrect. 
We experimented with the classical supervised classifiers (support vector machines and logistic regression) models, pre-trained language models (BERT$_{\text{Base}}$, BERT$_{\text{Large}}$, RoBERTa$_{\text{Base}}$, and RoBERTa$_{\text{Large}}$), as well as LLMs (Qwen3-8B, Qwen3-14B, Llama-3-8B-Instruct, Llama-3.3-70B-Instruct, Mistral-7B-Instruct-v0.3). To capture the granular information present in the documents provided as evidence, we split a document into multiple short documents by considering a window size of $n=3$. We tested the model for each short document. If the model predicted the label \textit{correct} for at least one such short document, we considered the prediction \textit{correct} for the document. To train the model, we consider the human-annotated span from the document and the answer sentence as a \textit{correct} pair. We computed the BM25 scores between the question and the fragmented documents and selected the top-scoring fragment as an incorrect pair, unless the fragment was part of the human-annotated span. 

\paragraph{Semantic Similarity and NLI Approach} To evaluate answer correctness, we also performed an in-depth analysis of the embedding model for encoding documents and human-annotated evidence, using it in two settings: cosine similarity and NLI-based evaluation. We performed the experiments with three embedding models: all-MiniLM-L6-v2, all-mpnet-base-v2, and sup-simcse-roberta-large. We aim to analyze the similarity between the document and the answer sentence representation compared to the NLI model, which takes the document and the answer sentence as input and produces labels such as \textit{supports}, \textit{refutes}, and \textit{insufficient information}. To achieve this, we experimented on a \test{}  dataset where the answer-supported documents and evidence to support the answers are annotated by experts. The answer could have been marked as supported by considering the continuous sentences from the document; therefore, we grouped the three sentences\footnote {We experimented with grouping 1, 2, 3, and 4 sentences and found that three is the most optimal on the \val{} dataset} for document-answer experiments. We extracted the sentence embeddings for the grouped sentences and the answer sentence, assessed the cosine similarity score, and also used the NLI model\footnote{\url{https://huggingface.co/FacebookAI/roberta-large-mnli}} to evaluate the model's probability of making the document-answer pair as support. For a comparative analysis, we also sampled a negative question from the dataset. We performed cosine similarity and NLI computations, using the same answer sentence as we did for the positive sample. We have outlined the experimental pipeline in the Algorithm \ref{algo:answer-correctness-analysis}.

\subsection{Citation Evaluation} \label{sec:citation_eval}
In addition to our answer evaluation, we assess the ability of language models to attribute claims to their cited sources. Specifically, we focus on the task of natural language inference, where models must determine whether a given citation supports, contradicts, or is neutral with respect to a target claim. The system-generated answers contain identifiers (PMID) of the citations, corresponding to the PubMed abstract of a scientific article that should support the claims made in that sentence. For each of these citations, medical informatics experts have labeled the sentence-document pair as 'supporting', 'contradicting', 'neutral', or 'not relevant'. We evaluate three different settings for precision and recall.

\subsubsection{Answer Sentence-Document}
Our first setting evaluates the ability of large language models to attribute a single sentence to the title and abstract of the cited PubMed document. We use a variety of model types, including both base and fine-tuned LLMs and sequence-to-sequence (seq2seq) models. Some models are not capable of providing ternary labels; we, therefore, evaluate the models in binary and ternary classification settings. For binary classification, we task the model to label the answer sentence and document pair as 'attributable' or 'not attributable' using the prompt in figure \ref{fig:pr:binary}. We then map 'supporting' labels in our dataset to 'attributable' and all other labels to 'not attributable'. For ternary classification, we task the model with labeling the answer sentence and document pair as 'supporting', 'contradicting', or 'neutral' using the prompt in figure \ref{fig:pr:ternary}.

\subsubsection{Answer Sentence-maxSimSentence Document}
In the second setting, we preprocess the PubMed abstracts to find the sentence most similar to the answer. We use the NLTK implementation of the Punkt sentence tokenizer to split each abstract into a list of sentences~\cite{10.5555/1717171, kiss-strunk-2006-unsupervised} We then use \minilmmodel{}\footnote{\url{https://huggingface.co/sentence-transformers/all-MiniLM-L6-v2}} to produce sentence embeddings for each sentence of the abstract. The cosine similarity of the answer sentence and each document sentence is calculated. We call the document sentence with the greatest cosine similarity to the answer sentence, maxSimSentence. The task is the same as the first setting, except only the maxSimSentence is given, rather than the entire abstract.

\subsubsection{Answer Nuggets-Document Nuggets}
In the third setting, we first process the answers and documents into lists of nuggets (atomic facts). This step is done by prompting Llama 3.3 with the instructions found in figure \ref{fig:pr:nugget_generation}. The models are then given the entire lists of nuggets from the answer and from the document and asked to assign one of four labels: 'supports', 'contradicts', 'neutral', or 'not relevant'. If at least one nugget generated from the answer is supported by at least one nugget from the document, that answer-document pair is labeled 'supports', unless any of the document nuggets contradict any of the answer nuggets. If at least one contradiction is detected, the answer-document pair is labeled 'contradicts'. If none of the answer nuggets are contradicted, and at least one of the answer nuggets is not supported, this answer-document pair is 'neutral'. Finally, if the document nuggets are unrelated to the answer nuggets, the answer-document pair is labeled 'not relevant'. This prompt was selected because it is closely aligned with instructions given to the human annotators for the creation of the \test{} dataset.




\section{Data  Availability}
All the experiments are performed on the \test{}, which is publicly available at the Open Science Framework repository (\url{https://osf.io/ydbzq/}). 

\section{Code Availability}
The \bioace{} code is available at GitHub (\url{https://github.com/deepaknlp/BioACE/}).

\section{Acknowledgments}
This research was supported by the Intramural Research Program of the National Institutes of Health (NIH). The contributions of the NIH authors are considered Works of the United States Government. The findings and conclusions presented in this paper are those of the authors and do not necessarily reflect the views of the NIH or the U.S. Department of Health and Human Services.
 \section{Author Contributions }
D.G. and D.D.F. conceived the study.   D.G., D.B., and D.D.F.  drafted the manuscript and analyzed the results.
D.G. and  D.B. developed the evaluation tools and performed the experiments. All authors reviewed the manuscript.
\section{Competing interests}
There are no competing interests. 
\bibliography{sn-bibliography}

\newpage
\begin{appendices}

\section{Results}\label{secA1}

\begin{table}[h]
\centering

\resizebox{\columnwidth}{!}{%
\begin{tabular}{lllll}
\hline
\multicolumn{1}{c}{\multirow{2}{*}{\textbf{\begin{tabular}[c]{@{}c@{}}Embedding\\ Model\end{tabular}}}} &
  \multicolumn{2}{l}{\textbf{Recall (R+B)--SimCSE}} &
  \multicolumn{2}{l}{\textbf{Recall (R+B)--ST}} \\
\multicolumn{1}{c}{} &
  \multicolumn{1}{l}{\textbf{Spearman}} &
  \textbf{Kendall} &
  \multicolumn{1}{l}{\textbf{Spearman}} &
  \textbf{Kendall} \\ \hline \hline
all-MiniLM-L6-v2         & \multicolumn{1}{l|}{0.908} & 0.758 & \multicolumn{1}{l|}{0.876} & 0.724 \\ 
all-mpnet-base-v2        & \multicolumn{1}{l|}{0.879} & 0.721 & \multicolumn{1}{l|}{0.846} & 0.683 \\ 
sup-simcse-roberta-large & \multicolumn{1}{l|}{0.876} & 0.712 & \multicolumn{1}{l|}{0.823} & 0.651 \\ \hline \hline
\end{tabular}%
}
\caption{Correlation coefficients in the system ranking considering the nugget-based recall metric vs recall metric computed using the human annotation of the machine-generated answers \cite{gupta2025dataset,gupta2024overview}. R+B refers to the setting \textbf{Required and Borderline} where answer sentences deemed as either required or borderline were included for clustering to compute the recall.}
\label{tab:res-prec_recall_correlation}
\end{table}

\begin{figure}[ht]
\centering
\begin{promptbox}{Completeness Prompt}\label{pr:completeness}
\small
You are an expert annotator. Given a question and an answer sentence, your task is to assign a single label from the following list: [`Required', `Unnecessary', `Borderline', `Inappropriate'].     The label definitions are as follows: \\
    \textbf{Required}: The answer sentence is necessary to have in the generated answer for completeness of the answers.\\
    \textbf{Unnecessary}: The answer sentence is not required to be included in the generated answer. An answer sentence may be unnecessary for several reasons:
    \begin{enumerate}
        \item  If including it would cause information overload, if it is added to the answer;
       \item If it is trivial, e.g., stating that many treatment options exist.
    \item If it consists entirely of a recommendation to see a health professional.
    \item If it is not relevant to the answer, e.g., describing the causes of a disease when the question is about treatments,
    \end{enumerate}
    
    \textbf{Borderline}: If an answer sentence is relevant, possibly even “good to know,” but not required, the answer sentence may be marked borderline.\\
    \textbf{Inappropriate}: The assertion may harm the patient, e.g., if, according to the answer, physical therapy reduces the pain level, but the patient experiences more pain due to hip mobilization, the patient may start doubting they are receiving adequate treatment.
    Do not generate anything else. 
    Respond ONLY with the label, no explanation.\\
    \textbf{Question} : \{question\} \\
    \textbf{Answer Sentence}: \{answer sentence\}
\end{promptbox}
\caption{Prompt used for answer completeness evaluation.}
\label{fig:pr:completeness}
\end{figure}

\begin{figure}[ht]
\centering
\begin{promptbox}{Binary Label Prompt}\label{pr:binary}
\small
\textbf{\#\#\# Instruction}:
Please solely verify whether the reference can support the claim. Options: 'attributable' or 'not attributable'. \\

\textbf{\#\#\#Input}:

\textbf{Claim}: \{sentence\}

\textbf{Reference}: \{document\} \\

\textbf{\#\#\# Output}:
\end{promptbox}
\caption{Prompt used for binary labeling of citations.}
\label{fig:pr:binary}
\end{figure}

\begin{figure}[ht]
\centering
\begin{promptbox}{Ternary Label Prompt}\label{pr:ternary}
\small
\textbf{\#\#\# Instruction}:
Please solely verify whether the reference can support the claim. Options: 'support', 'contradict', or 'neutral'. \\

\textbf{\#\#\# Input}:

\textbf{Claim}: \{sentence\}

\textbf{Reference}: \{document\} \\

\textbf{\#\#\# Output}:
\end{promptbox}
\caption{Prompt used for ternary labeling of citations.}
\label{fig:pr:ternary}
\end{figure}

\begin{figure}[ht]
\centering
\begin{promptbox}{Nugget Label Prompt}
\small
For the following lists of answer and document nuggets, select one of the following labels: \\

\textbf{Supports}: There is at least one document nugget that supports/agrees with each answer nugget.

\textbf{Contradicts}: There is at least one document nugget that disagrees with an answer nugget or states its opposite.

\textbf{Neutral}: The document nuggets are topically relevant, but lack any information to validate or invalidate the all of the answer nuggets.

\textbf{Not relevant}: The document nuggets are not relevant to the answer nuggets. \\

\textbf{Answer Nuggets}: \{list of answer nuggets\}

\textbf{Document Nuggets}: \{list of document nuggets\}
\end{promptbox}
\caption{Prompt used for labeling answer and document nugget lists.}
\label{fig:pr:nugget}
\end{figure}

\begin{figure}[ht]
\centering
\begin{promptbox}{Nugget Generation Prompt}
\small
List all of the information nuggets in the text given below. Each nugget must contain one, and only one, fact from the text. A nugget must be as concise and as specific as possible. Each element in a list must be its own nugget. The list of nuggets must not contain redundant information. Return a list of nuggets such that each nugget is on a new line. Do not number or bullet the list. Do not include anything in your response except for the list of nuggets. Here is an example of the output format: \\

nugget1 \\
nugget2 \\
… \\

Here is an example text: During infections, a battle for iron takes place between the human host and the invading pathogens. Lymphocytes need iron to mount an effective cellular and humoral response. Viruses depend on iron to replicate within living host cells. During the acute phase of infection, blood levels of iron decrease. Ferritin levels are high. Elevated serum ferritin is associated with increased mortality. As a major iron storage protein, ferritin is essential to iron homeostasis and is involved in a wide range of physiologic and pathologic processes. The inflammation cascade and poor prognosis of COVID-19 may be attributed to high ferritin levels. Iron depletion therapy was proposed as a novel therapeutic approach in the COVID-19 pandemic. \\

This is the list of nuggets that should be extracted from this text: \\

Lymphocytes and viruses compete for iron. \\
Lymphocytes need iron for cellular response. \\
Lymphocytes need iron for humoral response. \\
Viruses need iron to replicate. \\
Infection lowers iron levels in the blood. \\
Infection increases ferritin levels in the blood. \\
High ferritin is associated with increased mortality. \\
Iron homeostasis needs ferritin. \\
Ferritin is involved in physiologic processes. \\
Ferritin is involved in pathologic processes. \\
High ferritin indicates response to inflammation. \\
High ferritin levels are linked to poor outcomes of COVID-19. \\
Iron depletion therapy showed anti-viral activity in the COVID-19 pandemic. \\
Iron depletion therapy showed anti-fibrotic activity in the COVID-19 pandemic. \\

Text: \{text\}
\end{promptbox}
\caption{Prompt used for generating nuggets from text.}
\label{fig:pr:nugget_generation}
\end{figure}


\begin{table}[]
\centering
\resizebox{\columnwidth}{!}{%
\begin{tabular}{llccc}
\hline
\textbf{Model} & \textbf{Setting} & \textbf{Precision} & \textbf{Redundancy} & \textbf{Harmfulness} \\ \hline \hline
BERT$_{\text{Base}}$                   & Fine-tuned & 0.834/0.669 & 0.860/0.698 & -/- \\ 
BERT$_{\text{Large}}$      & Fine-tuned & 0.849/0.692 & 0.818/0.615 & -/- \\ 
RoBERTa$_{\text{Base}}$            & Fine-tuned & 0.749/0.563    & 0.849/0.634  & -/- \\ 
RoBERTa$_{\text{Large}}$           & Fine-tuned & 0.827/0.664  & 0.894/0.726 & -/- \\ \hline
Llama-3.3-70B-Instruct   & Zero-shot  & 0.859/0.715 & 0.816/0.629 & 0.272/0.226  \\ 
Llama-3.3-70B-Instruct   & Fine-tuned & 0.721/0.531 & 0.634/0.470 & 0.411/0.369  \\ 
Llama-3-8B-Instruct & Zero-shot  & 0.740/0.605 & 0.440/0.332 & 0.091/0.102 \\ 
Llama-3-8B-Instruct & Fine-tuned &  0.754/0.603 & 0.616/0.484 & -/- \\
Qwen3-14B                      & Zero-shot  & 0.613/0.460 & 0.133/0.106 & 0.147/0.133  \\ 
Qwen3-14B                     & Fine-tuned & 0.724/0.549 & 0.469/0.342 & 0.219 \\ 
Qwen3-8B                      & Zero-shot  & 0.853/0.686 & 0.047/0.038 & 0.283/0.228 \\ 
Qwen3-8B                       & Fine-tuned & 0.541/0.408 & -0.186/-0.139 & 0.195/0.177 \\ 
Mistral-7B-Instruct-v0.3  & Zero-shot  & 0.813/0.658 & -/- & -/- \\ 
Mistral-7B-Instruct-v0.3  & Fine-tuned & 0.705/0.554 & -0.139/-0.116 & -/- \\ \hline \hline
\end{tabular}%
}
\caption{Correlation coefficients (Spearman/Kendall Tau) in the system ranking considering the model-predicted answer sentences for assessing its relevance vs relevance metric computed using the human annotation of the machine-generated answers \cite{gupta2025dataset,gupta2024overview}. The -/- indicates the undefined coefficients where systems could not predict any \textit{Unnecessary} and \ \textit{Inappropriate} labels, resulting in zero Redundancy and Harmfulness scores and the same ranks for all systems.}
\label{tab:res-completeness_correlations}
\end{table}

\begin{table}[]
\centering
\resizebox{0.9\columnwidth}{!}{%
\begin{tabular}{lllll}
\hline
\textbf{Model} & \textbf{Setting} & \textbf{Precision} & \textbf{Recall} & \textbf{F1-Score} \\ \hline \hline
Llama-3.3           & Base & 76.10 & 77.50 & 76.79 \\
FLAN-T5             & Base & 74.80 & 77.20 & 75.98 \\
FLAN-UL2            & Base & 75.20 & 77.80 & 76.48 \\
Llama-3.3           & Fine-tuned & 77.90 & 78.50 & 78.20 \\
FLAN-T5             & Fine-tuned & 76.90 & 77.50 & 77.20 \\
FLAN-UL2            & Fine-tuned & 74.50 & 78.90 & 76.63 \\ \hline \hline
\end{tabular}%
}
\caption{Performance of base and fine-tuned models when tasked with assigning ternary labels to a claim sentence and the sentence with the highest cosine similarity to the claim.}
\label{tab:res-ternary-maxSimSentence}
\end{table}
\begin{table}[h]
\centering
\resizebox{\columnwidth}{!}{%
\begin{tabular}{lllllll}
\hline
\multicolumn{1}{c}{\multirow{2}{*}{\textbf{Model}}} &
  \multicolumn{2}{l}{\textbf{Citation Coverage}} &
  \multicolumn{2}{l}{\textbf{Citation Support Rate}} &
  \multicolumn{2}{l}{\textbf{Citation Contradict Rate}} \\ 
\multicolumn{1}{c}{} &
  \textbf{Spearman} &
  \textbf{Kendall} &
  \textbf{Spearman} &
  \textbf{Kendall} &
  \textbf{Spearman} &
  \textbf{Kendall} \\ \hline \hline
Answer Sentence-Document & 0.961 & 0.847 & 0.383 & 0.276 & 0.248 & 0.205 \\
Answer Sentence-maxSimSentence Document & 0.960 & 0.847 & 0.389 & 0.281 & 0.248 & 0.205 \\
Answer Nuggets-Document Nuggets & 0.975 & 0.884 & 0.578 & 0.391 & 0.396 & 0.309 \\
\hline \hline
\end{tabular}
}
\caption{Correlation coefficients comparing rankings based on automatic and manual evaluation of \test{}.}
\label{tab:res-citation_correlation}
\end{table}

\begin{table}[]
\centering
\resizebox{0.9\columnwidth}{!}{%
\begin{tabular}{lllll}
\hline
\textbf{Model} & \textbf{Setting} & \textbf{Precision} & \textbf{Recall} & \textbf{F1-Score} \\ \hline \hline
Llama-3.3           & Base  & 75.90 & 77.60 & 76.74 \\
FLAN-T5             & Base  & 74.60 & 77.10 & 75.83 \\
FLAN-UL2            & Base  & 75.40 & 77.50 & 76.44 \\
Llama-3.3           & Fine-tuned & 76.80 & 76.30 & 76.54 \\
FLAN-T5             & Fine-tuned & 76.70 & 77.60 & 77.15 \\
FLAN-UL2            & Fine-tuned & 74.70 & 78.80 & 76.69 \\ \hline \hline 
\end{tabular}%
}
\caption{Performance of base and fine-tuned models when tasked with assigning ternary labels to a the list of nuggets contained in the claim and the list of nuggets contained in the PubMed abstract.}
\label{tab:res-ternary-nuggets}
\end{table}

\begin{table}[]
\centering
\resizebox{0.9\columnwidth}{!}{%
\begin{tabular}{lllll}
\hline
\textbf{Model} & \textbf{Setting} & \textbf{Precision} & \textbf{Recall} & \textbf{F1-Score} \\ \hline \hline
Llama-3.3           & Base        & 75.34 & 76.21 & 75.77 \\
FLAN-T5             & Base        & 74.09 & 76.77 & 75.41 \\
FLAN-UL2            & Base        & 74.66 & 76.88 & 75.75 \\
Llama-3.3           & Fine-tuned  & 77.42 & 77.91 & 77.66 \\
FLAN-T5             & Fine-tuned  & 76.55 & 77.06 & 76.80 \\
FLAN-UL2            & Fine-tuned  & 73.87 & 78.69 & 76.21 \\ \hline \hline
\end{tabular}%
}
\caption{Performance of base and fine-tuned models when tasked with assigning ternary labels to a claim sentence and a PubMed title and abstract.}
\label{tab:res-ternary}
\end{table}
\begin{algorithm}[!h]
\small
\SetAlgoLined
\SetNlSty{textbf}{}{.} 
\SetNlSkip{1em}    
\SetKwInput{KwIn}{Input}
\SetKwInput{KwOut}{Output}

\KwIn{Dataset $\mathcal{D} = \{D_1, \dots, D_N\}$ with questions and answer sentences, sentence embedding model $\mathcal{E}$, NLI model $\mathcal{N}$}
\KwOut{Overall evaluation metrics $\mathcal{M}$}

\vspace{0.3em}

Initialize per-question result map $\mathcal{R} \gets \emptyset$\;

\ForEach{question-answer sentences $D_i \in \mathcal{D}$}{
    Let question $q \gets$ $D_i[question]$\;
    \ForEach{answer sentence $S_j$ in $D_i$}{
        
         \Comment{\textcolor{teal}{\textit{Cosine similarity and NLI computation with positive sample}}} \;
        \Indp
        Extract supporting documents $S_j[suppoting]$ \;
        Groped three sentences together and created a  supporting documents list $S_j^{supp}$\;
        Computed cosine similarity between the embedding of each item of the supporting documents list and the answer sentence $S_j$ using $\mathcal{E}$\; 
        Get the maximum cosine similarity and assign it as the document-answer sentence \texttt{Sim(doc, answer sent) }\;

        Computed NLI scores between each item of the supporting documents list and the answer sentence $S_j$ using $\mathcal{N}$\; 
        Get the maximum NLI score of the \textit{Supports} label and assign it as the document-answer sentence \texttt{NLI(doc, answer sent) }\;

        Repeat steps 5-10 with the human-annotated evidence in the supporting documents instead of groping and computing the scores for all the groped sentences. This step yields \texttt{Sim(evidence, answer sent) } and \texttt{NLI(evidence, answer sent) }.
       
          \Comment{\textcolor{teal}{\textit{Cosine similarity and NLI computation with negative sample}}} \;

        Sample a different row $D_k, k \neq i$\ and pick the first answer sentence $S_k$\;
        Repeat steps 6-12 with the $S_k$ and the answer sentence  $S_j$. \;
    
        Add scores to $\mathcal{R}$.
    }
}
\Comment{\textcolor{teal}{\textit{Aggregate scores across all answer sentences}}}\;
\ForEach{question $q$ in $\mathcal{R}$}{
Average cosine similarity and NLI scores for positive and negative samples \;
Compute Accuracy as a fraction of positive scores $>$ negative scores \;
Compute AUC with positive and negative scores\;
}

\BlankLine
  
$\mathcal{M} \gets \text{Aggregate}(\{\mathcal{R}_q\})$ \Comment{\textcolor{teal}{\textit{Aggregate metrics across all questions}}}\;

\Return $\mathcal{M}$\;
\vspace{0.3em}
\caption{Answer correctness evaluation algorithm to analyze the embedding model, NLI model, and cosine similarity scores between document/evidence and answer sentence.}
\label{algo:answer-correctness-analysis}
\end{algorithm}

\begin{figure}[ht]
    \centering
    \begin{subfigure}[b]{0.49\textwidth}
        \centering
        \includegraphics[width=\textwidth]{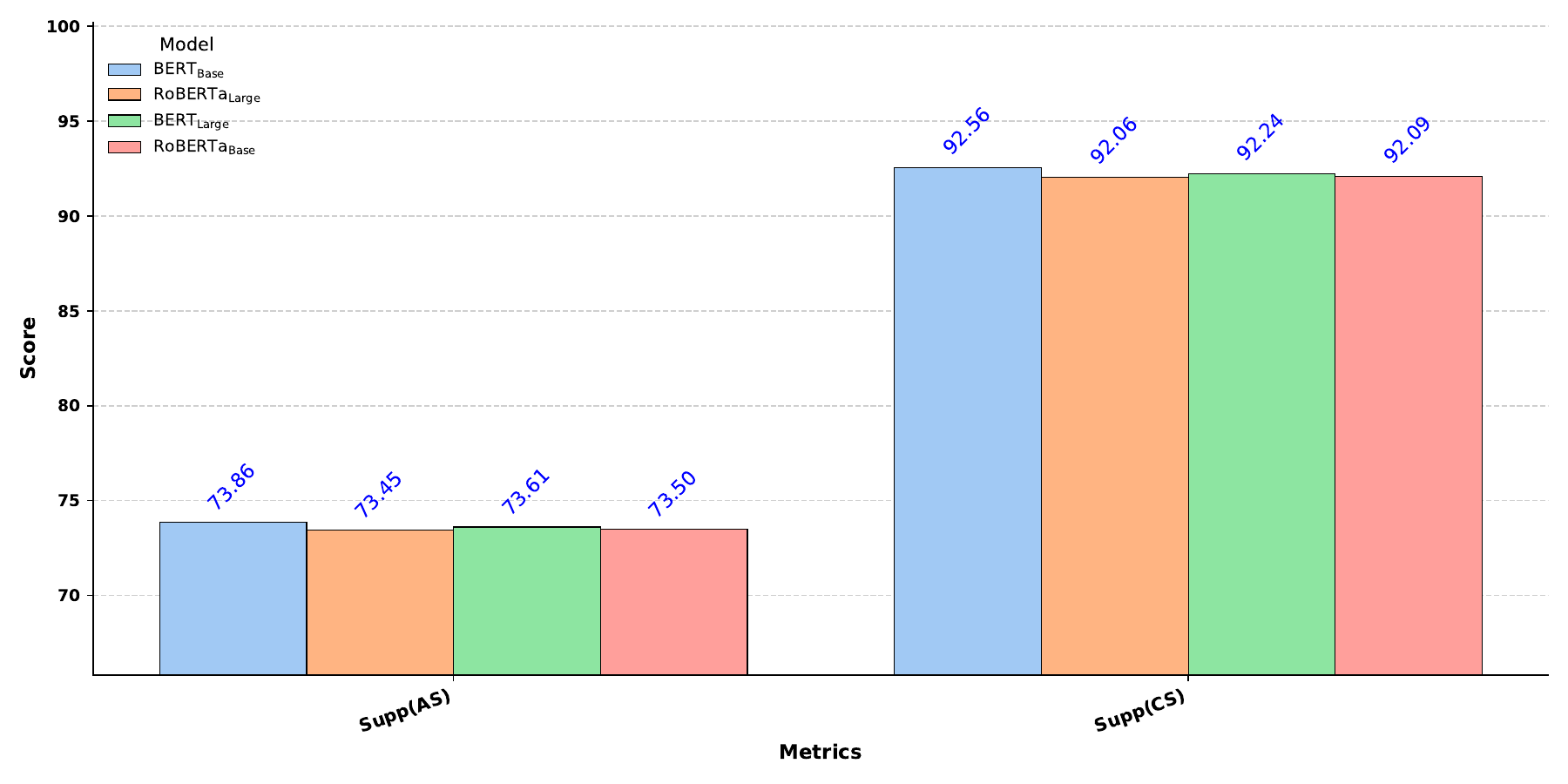}
        \caption{}
        \label{fig:ac-participant-analysis}
    \end{subfigure}
    \hfill
    \begin{subfigure}[b]{0.49\textwidth}
        \centering
        \includegraphics[width=\textwidth]{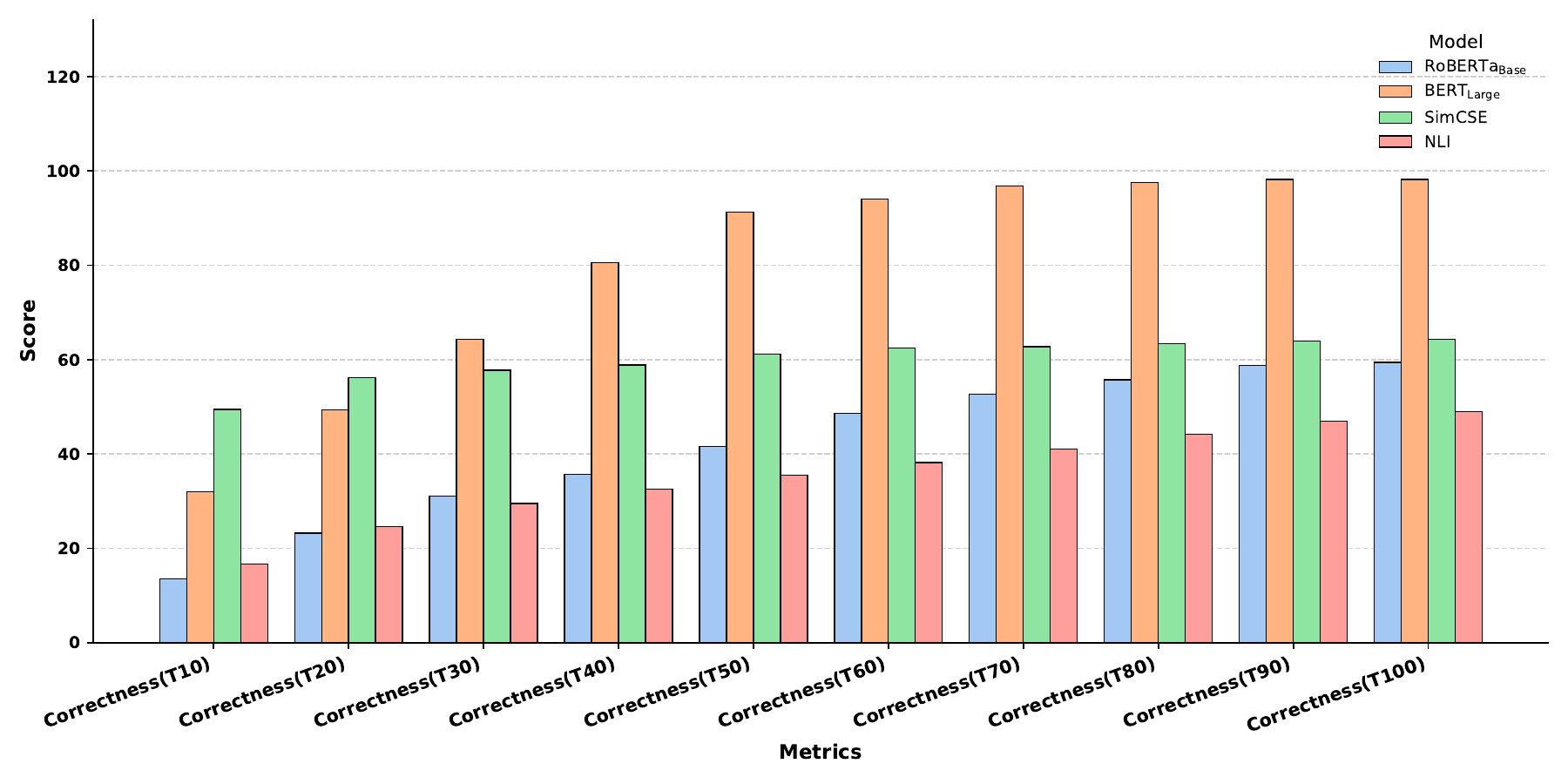}
        \caption{}
        \label{fig:ac-top-doc-analysis}
    \end{subfigure}
    
    \caption{Answer correctness results using different PLMs: \textbf{(a) } comparing the effectiveness of the model in terms of human-assessed supported documents and all model-predicted documents, \textbf{(b)} comparing the effectiveness of the model with the top-k retrieved documents. }
    \label{fig:ac-participant-top-doc}
\end{figure}

\end{appendices}

\end{document}